\def\year{2015}
\documentclass[letterpaper]{article}
\usepackage{aaai}
\usepackage{times}
\usepackage{helvet}
\usepackage{courier}

\usepackage{times}
\usepackage{url}
\usepackage{latexsym}
\usepackage{graphicx}
\usepackage{indentfirst}
\usepackage{setspace}
\graphicspath{ {figures/}{figures/eps/}{figures/pdf/} {figures/png/}}% specify the path where figures are located
\usepackage{multirow}
\usepackage{algorithm, algorithmic}
\usepackage[tableposition=top]{caption}
\usepackage[T1]{fontenc}
\usepackage{natbib}
\usepackage{tikz}

\newcommand{\ignore}[1]{}
\begin{document}
\title{A Data-Driven Approach for Semantic Role Labeling from Induced Grammar Structures in Language}
\author{
Vivek Datla$^{1}$,  David Lin$^{2}$, Max Louwerse$^{3}$,\\
\Large\bf Abhinav Vishnu$^{4}$\\
$^{1}$Philips Research, Cambridge, MA;
$^{2}$Baylor University, Waco, TX;\\
$^{3}$Tilburg University, Tilburg, Netherlands;
$^{4}$Pacific Northwest National Laboratory, Richland, WA\\
vivek.datla@philips.com, davidlin@baylor.edu, \\maxlouwerse@tilburg.edu, abhinav.vishnu@pnnl.gov\\
}
\maketitle

\begin{abstract}
 Semantic roles play an important role in extracting knowledge from text. Current unsupervised approaches utilize features from grammar structures, to induce semantic roles. The dependence on these grammars, however, makes it difficult to adapt to noisy and new languages. In this paper we develop a data-driven approach to identifying semantic roles, the approach is entirely unsupervised up to the point where rules need to be learned to identify the position the semantic role occurs. Specifically we develop a modified-ADIOS algorithm based on ADIOS \cite{solan2005} to learn grammar structures, and use these grammar structures to learn the rules for identifying the semantic roles based on the context in which the grammar structures appeared. The results obtained are comparable with the current state-of-art models that are inherently dependent on human annotated data.
 
 \end{abstract}

 \section{Introduction}
For speakers and hearers to understand the meaning of the message, they need to understand the ``who-did-what-to-whom'', the semantic roles within a sentence. These semantic roles are used to denote the underlying relationship between various constituents of a sentence, mostly with the main predicate (verb) of the sentence. For example, in the sentence \emph{``John kissed Mary'',} we denote \emph{``John''} as the ``Agent'' and \emph{``Mary''} as the ``Patient'' with respect to the relation to main verb \emph{``kissed''}. 

Semantic roles play a crucial role in a variety of NLP applications because they provide crucial information to the meaning of the individual words. The role of ``Mary'' is different when Mary is a Patient (``John kissed Mary'') or an Agent (``Mary kissed John'').
 Note that the syntactic Subject or Object might be different while the Agent and Patient may remain the same (e.g., ``John kissed Mary'' and ``Mary was kissed by John''). NLP applications in which semantic roles are frequently used include information extraction \ignore{\cite{surdeanu2003}}, question answering \ignore{\cite{frank2007,narayanan2004}}, automatic summarization \ignore{\cite{melli2005}}, and many more. \ignore{\cite{surdeanu2003} used the predicate-argument structure ( a structure that encodes the semantic role information) to build templates for robust information retrieval. \cite{frank2007} identified that questions such as \emph{``what''} could be answered by taking into account the semantic roles in which the answer occurs in that particular domain. Thus, developing effective automatic semantic role labeling(SRL) algorithms can be very beneficial.}

One can view SRL as a two-step process: 1) Identifying the slots in a sentence that are most likely to contain the semantic roles, and 2) Selecting a specific semantic role that mostly likely fits those slots. Computationally both these problems are framed as machine learning models.

While there has been a lot of work on supervised SRL methods, supervision does have its limitations: 
\begin{itemize}
\item Annotating gold standard data is a slow and resource intensive process
\item Dealing with novel words or novel word combinations can create problems (``I googled about pizza today''); current models cannot handle mixed languages, making it easy for existing models to become obsolete
\ item  SRL models trained in one domain usually do not perform well in other domains \cite{pradhan2005b}.
\end{itemize}
Exploring the effectiveness of using unsupervised learning methods for SRL is, therefore, a worthwhile exercise.

Several unsupervised approaches have been developed to extract semantic roles from text (see Table \ref{table:literature_survey}). These studies apply methodologies to exploit the structure of the language to extract semantic roles. The structures are built using syntactic parsers built from large human annotated corpora. Similar semantic roles tend to appear in the same part of the dependency tree, and unsupervised semantic role labeling models exploit this phenomenon. However, even though these methods do not need a large corpus for role labeling, they do require (large) human annotated data for building the structures (e.g. dependency tree, part-of-speech and constituent labels) for the labeling task. The current methods for unsupervised semantic role labeling still suffer from the limitations above. Table \ref{table:literature_survey} indicates the supervised, unsupervised and semi-supervised components in various parts of semantic role labeling. 

\begin{table}
\centering
\caption{Human annotated data dependency among several components of unsupervised SRL systems; part-of-speech(1); parsers(2); slot for semantic role(3); labeling semantic role(4); Yes(Y), No(N), partial(Y$_p$)}
 \begin{tabular}{l l l l l}
 \hline
Literature&1&2&3&4\\ \hline
\cite{abend2009}&Y&Y&Y&N \\ 
\cite{garg2012}&Y&Y&Y&N \\ 
\cite{grenager2006}&Y&Y&Y$_p$&Y$_p$\\
\cite{lang2010}&Y&Y&Y&N\\
\cite{lang2011a}&Y&Y&Y&N\\
\cite{modi2012}&Y&Y&Y&N\\
\cite{titov2011}&Y&Y&Y&N\\
\textbf{our method}&\textbf{N}&\textbf{N}&\textbf{Y$_p$}&\textbf{Y$_p$}
\\
\hline
\end{tabular}
\label{table:literature_survey}
\end{table}

The main goal of the current paper is to build a model that extracts semantic roles present in language using bottom-up or data-driven approaches, without the need of human annotated data for structure building. Our method starts out by learning patterns of the language using re-occurring phrases within a context. We learn more complex patterns by increasing the context on existing patterns, and we cluster the patterns into groups based on word overlap among patterns. Specialization increases the content inside patterns and generalization helps to find equivalent patterns in a given context. Since patterns are learned from language, and patterns are organized hierarchically, identifying the semantic role in one of the patterns would help us percolate the semantic information to all parents of the pattern. 
This organization of patterns gives us the additional advantage of needing only a relatively small training set of known semantic roles (e.g. proportional to the number of patterns) for an effective SRL model. 

\begin{figure}
 
 \centering
 \includegraphics[width=0.25\textwidth,natwidth=610,natheight=642]{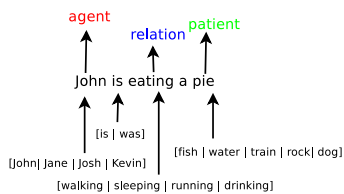}
  \caption{Thematic role of equivalent sentences}
  \label{fig:semanticRoleEquivalentSentence}
  
\end{figure}

Figure \ref{fig:semanticRoleEquivalentSentence} illustrates our method. Starting with a sentence ``John is eating a pie'', we build a pattern of ``X is Y a Z'' where X, Y, Z are slots that contain multiple terms that appear in sentences in our unannotated corpus that share the pattern.  Here we can see the location of the agent and patient on the relation in that structure does not change. This consistency in the structure allows us to learn the semantic roles based on the patterns. 

We modeled our pattern learning method based on a modified version of ADIOS algorithm \cite{solan2005}, one of the few algorithms that model the learning of language as bottom-up machine learning process. Following the description of the process how to create the patterns and how to use them for SRL process, we will present a computational study showing the success of our methods that do not require a large annotated corpus compared to alternative methods.  %It identifies a significant pattern, gets those patterns that are equivalent to the significant pattern, and builds and reuses the patterns learned to build new patterns.

\section{Data-driving SRL procedure}

Our proposed semantic role labeling model consists of two steps:

\begin{itemize}
    \item Extracting patterns and rules from the text: The first step is to take a {\em unannotated} corpus of text and discover common patterns and structures embedded in them. As mentioned, a pattern is not only a recurrence of statements but also adherence to a form. For example ``X Y a Z'' represents a form where the letters denote set of terms that can appear in that location. The terms may be a single word (e.g. X=John, Y=eats, Z=pie) or they can also be other patterns (e.g. Z = U made in V, where U=pie, V=France). Thus, the patterns are inherently organized to form hierarchical structures.
    Once we extract the patterns and structures, they are transformed into (production) rules. These rules are then used later for the parsing of the text in the SRL task.
    
    \item Semantic role learning/labeling: We use the rules learned to parse sentences of a given training set, generate features from the parsed sentences and use those features to train a classifier for identifying semantic roles. We use a small set of human annotated data to train the classifier.
\end{itemize}

The following sections present these steps in detail.

\subsection{Extracting patterns and rules via Modified ADIOS(m-ADIOS)} \label{sec:mADIOS}

To extract patterns and structures from the text, we apply a modified version of the Automatic Distillation of Structure (ADIOS) algorithm \cite{solan2005}. The algorithm builds generative rules based on the symbolic sequential information (i.e. language input as text).

In the ADIOS algorithm, a corpus is represented by a directed multigraph, where each vertex denotes a word. Each sentence is represented by a path connecting these words. If a phrase (e.g. John is) appears in multiple sentences, each occurrence will contribute a separate edge between the words (hence the multi-graph). Each sentence is complemented by a ``begin'' and an ``end'' vertex. The multigraph representation helps to identify those vertices that are thickly connected. They inherently represent the sequences that frequently appear in the corpus.

The algorithm iteratively picks a path (say $v_1, \ldots, v_n$), and determines the part of the sentence that qualifies as a significant pattern. For each term ($v_i$) in the path, we calculate the conditional probability that a path from $v_1$ to $v_{i-1}$ will continue to $v_i$ (denoted as $P_R(v_1, v_i)$), and use a statistical test to detect the most significant drop in $P_R(v_1, v_i)$. This determines the right end of the pattern. A similar construct $P_L(v_n, v_i)$ -- the conditional probability that sentences go from $v_{n}$ to $v_{i+1}$ would have come from $v_i$ is used to find the left end of the pattern. Once the significant pattern is found, it is replaced by a single vertex and the graph is rewired.

Note that the new vertex does not replace the vertices in the path, but only consolidates all the paths represented by the pattern into a vertex -- for example, if $v_2, v_3, v_4$ forms a significant pattern, then a vertex $P$ is used to represent all paths that pass through the three vertices. $v_2, v_3, v_4$ are not removed, if there are other alternate routes between $v_2, v_3 , v_4$ and other vertices.
The single vertex that represents the pattern can also be looked at as a hierarchical structure. These hierarchical structures (new vertices) appear in a context. Consequently, we search through all the paths in which this vertices appear and select a significant path and generalize on the vertices that appear in the same context.

Furthermore, a generalization process of the patterns allows for consolidation of the patterns. If two patterns differ by only one vertex, we put all vertices into one class called the Equivalence class and replace all the vertices inside the equivalence class by a single vertex.

These steps are repeated until all the paths are iterated and no other significant patterns are formed. The output of this process consists of rules that are similar to context-free grammar(CFG) rules that will be used in subsequent steps. 

\begin{figure} 
    \centering
    \includegraphics[width=0.40\textwidth,natwidth=610,natheight=642]{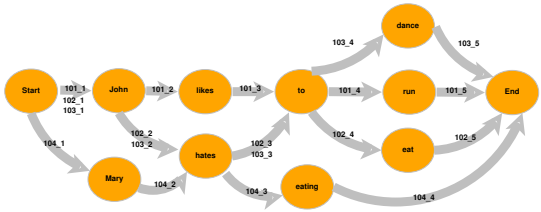}
    \caption{Data representation in modified-ADIOS}
    \label{fig:directedGraph}
\end{figure}

Some modifications are made to the ADIOS algorithm. First we consolidate the representation of the graph by using a simple directed graph, but with edge labels denoting the various sentence corresponds to each edge. This modification in the structure makes the graph scalable. For example, Figure \ref{fig:directedGraph} shows the graph built on a corpus of four sentences, \emph{John likes to run,} \emph{John hates to eat,} \emph{John hates to dance,} and \emph{Mary hates eating}.

The ADIOS model generalizes greedily and allows for structures with variable sizes to be represented as one structure. This leads to phrases of various sizes not occurring in the same context to be considered as one structure, there by creating unwanted ambiguity in the grammar. Our goal here is not to produce a very compact grammar but a non-ambiguous grammar. We, therefore, modified the ADIOS algorithm to learn only the structures that occur within a context and have a very strict method for generalization. 

\begin{figure}
 \centering
 \includegraphics[width=0.20\textwidth,natwidth=610,natheight=642]{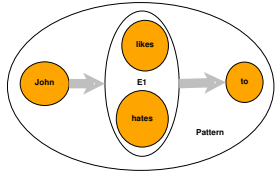}
  \caption{Pattern structure}
  \label{fig:Pattern}
\end{figure}

 We first create an equivalence class with all words that share the same left and right context. Then merge the equivalence class, and the respective left and right context to a single pattern. For example, in Figure \ref{fig:directedGraph} if ``John likes to'' is significant then in the context of ``John'' and ``to'' we find that ``likes'' and ``hates'' can be merged into one equivalence class. We merge both vertices and make an implicit note in our grammar tree that make the tokens``likes'' and ``hates'' fall into the same equivalence class. Vertices ``John'', ``equivalence class'' and ``to'' are then merged into a single vertex. The pattern learned from the text is shown in Figure \ref{fig:Pattern}.

It is important to note that we only replace these three vertices and two links and do not change any other vertices in the graph. 

After finding a significant pattern, we check if we can find equivalent middle words for the same left and right context. By finding the middle words, which share the same context, we are more likely to find words which if interchanged also form syntactically correct sentences. After finding the middle words that share the same left and right context, we put them into a group called equivalence class. 

Next, we check for the equivalence of a pattern. The equivalence of a pattern is defined as the maximal overlap of the pattern with the existing patterns. Since each pattern consists of a left context, a right context, and an equivalence class, we consider two patterns to be equivalent if two of the three components match. 

Every time we find a pattern we check if we can find an equivalence of that pattern to find structures that are replaceable. We repeat the process until we have covered all the paths and can no longer find any additional significant patterns. By following the above methodology, we end up with several equivalence classes that can be made up of patterns, words or other equivalence classes. A pattern can consist of words, equivalence classes and also other patterns. 

We also generalize the equivalent patterns if they share at least left or right context and have more than $\sigma$ overlap, i.e. $\frac{size( {E_{i} \bigcap {E_{j}}})} {{size( {E_{i} \bigcup {E_{j}}})}} \geq \sigma$.  The algorithm \ref{Algo:Generalization_Equivilence_classes} shows the steps followed for generalizing the equivalent classes. After finding the patterns, we consolidate all rules that we learned in the form of hierarchical structures. These hierarchical structures have the base patterns as leaf nodes, and patterns that use the base structures as parents of the base patterns and so on. As we move up the tree, we would be increasing the information in the pattern, as the size of the pattern keeps increasing as it subsumes the information from the nodes below. Even though the information increases, the nodes that are subsumed by this parent node would have the same dependency structure as the original base nodes. Since the structure is the same, identifying the semantic role slots in the base nodes would also be the slots for semantic roles in the parent nodes.

\begin{algorithm}
\caption{Build a sparse directed graph $G= (V, E)$ where each Sentence $s_i \in C$ is a path} 
\begin{algorithmic}
 \STATE Input: Sentences from Corpus $C$ 
\STATE Output: Sparse directed graph $G$ 
\end{algorithmic}
\label{Algo:BuildGraph}
\begin{algorithmic} 
\FOR { each sentence $s_i \in C$ }
\FOR {each word $w_n \in s_i$ }
\IF {$w_{n} \notin G $}
\STATE add $w_{n}$ to $G$
\ENDIF
\IF{ $w_{n+1} \notin G $}
\STATE add $w_{n+1}$ to $G$
\ENDIF
\IF{ An edge exists between $w_n$ and $w_{n+1}$ }
\STATE Add sentenceID and linkID to the edge list.
\ELSE
\STATE Add directed edge from $w_{n} \rightarrow w_{n+1}$ in Graph $G$
\STATE Add sentenceID and linkID to the edge list.
\ENDIF
\ENDFOR
\ENDFOR
\end{algorithmic}
\end{algorithm}

 \begin{algorithm}

\caption{Generalization of Equivalence Patterns}
\label{Algo:Generalization_Equivilence_classes}

\begin{algorithmic} 
\STATE $E_{i} = P_{i} \rightarrow {L_{i} E_{i} R_{i}}$
 \STATE $E_{j} = P_{j} \rightarrow {L_{j} E_{j} R_{j}}$
\IF {$(L_{i} = L_{j})$ AND Equivalence class overlap $(E_{i}, E_{j}) \geq \sigma $}
\STATE Build new Equivalence Class $E_{k} = R_{i} \bigcup R_{j}$
\STATE Merge Equivalence Patterns $E_{i}$ and $E_{j} \rightarrow E_{i,j}$
\STATE Build new Pattern $L_{i}$, $E_{i,j}$, $E_{k} \rightarrow P $
\ENDIF
\end{algorithmic}
\begin{algorithmic} 
\IF {$(R_{i} = R_{j})$ AND Equivalence class overlap $(E_{i}, E_{j} )\geq \sigma$}
\STATE Build new Equivalance Class $E_{k} = L_{i} \bigcup L_{j}$
\STATE Merge Equivalence Patterns $E_{i}$ and $E_{j} \rightarrow E_{i,j}$
\STATE Build new Pattern $E_{k}$, $E_{i,j}$, $R_{i} \rightarrow P$
\ENDIF
\end{algorithmic}
\begin{algorithmic} 
\IF{$(L_{i} = L_{j})$ AND $(R_{i}= R_{j} )$}
\STATE Merge Equivalence Patterns $E_{i}$ and $E_{j} \rightarrow E_{i,j}$
\STATE Build new Pattern $L_{i}$, $E_{i,j}$, $R_{i} \rightarrow P$
\ENDIF
\end{algorithmic}
\end{algorithm}

At the end of m-ADIOS, the patterns and classes take the form of a hierarchical structure. For us to use them in subsequent steps, we would derive rules from them, so that we can then parse any statement through the rules for the semantic role labeling task. Each rule contains a node of the tree on the left and all its children in sequential fashion on the right. 
Table \ref{table:cfgLikeRulesRawText} shows an example of CFG like rules created using patterns, equivalent classes, and reduced paths in the graph.

  \begin{table}
 
  \caption{CFG like rules generated using m-ADIOS}
  \begin{tabular}{l}
  \hline
 
 ROOT $\rightarrow$    S \\
 S $\rightarrow$    john Pattern\_118 mother\\ 
 Pattern\_118    $\rightarrow$ fedexed E\_56 his \\ 
 
 E\_56    $\rightarrow$ Pattern\_109 \\ 
 Pattern\_109 $\rightarrow$ his E\_55 to \\ 
 E\_55 $\rightarrow$computer | classes | heart | horse | opposition | \\socks | package | finger | cheeks | father | \ignore{\\hand | stand | future | yardwork | \\case | head | energy | ability} \\ \hline
  
  \end{tabular}
 
  \label{table:cfgLikeRulesRawText}
 
  \end{table}

\subsection{Semantic role labeling}

The semantic role labeling task involves learning from a corpus of sentences annotated with semantic roles. We treat each sentence as a single unit (as semantic roles relate terms from a sentence). We first parse each sentence, using the hierarchical structure rules learned from the m-ADIOS algorithm. After parsing the sentences, we locate patterns and collect contextual information around them. This information will be used as features for a classifier supporting learning and labeling purposes.

\subsubsection{Data: Propbank}
We illustrate this process by using PropBank \cite{kingsbury2002} as our corpus. PropBank annotated text from the Penn Treebank and the Wall Street Journal Corpus. PropBank is based on predicates, in the majority of the cases they are verbs. Each predicate has a set of arguments that is associated (from ARG0 to ARG5) associated to it. Arguments come with types such as location (LOC), temporal (TMP), manner (MNR), etc. The first two arguments, ARG0, and ARG1, are similar to prototypical agent and patient respectively. For the current purposes of the paper, we use only those sentences fewer than ten words. After filtering all $10600$ sentences, we end up with 2249 sentences. The 2249 sentences from PropBank includes $4892$ unique words which occur $20148$ times(on average each word occurs  $4.12$ times (SD$=51.05$)). In the current work, we concentrate only on the agent-patient-relation. Relation represents the verb for which agent and patient are semantic roles. 

\begin{figure}
 \centering
 \includegraphics[width=0.40\textwidth,natwidth=610,natheight=642]{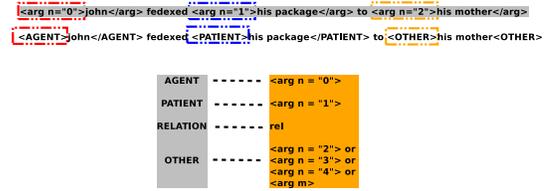}
 \caption{PropBank annotation scheme}
 \label{fig:PropBankTagsToCustomTags}
\end{figure}

Figure \ref{fig:PropBankTagsToCustomTags} shows an example of an annotated sentence by PropBank, and also simplified annotation for our learning task. <arg n =``0''> represents an Agent present in the sentence, we identify this information using the tag ``Agent'', <arg n=``1''> represents the Patient, we identify this information using the tag ``Patient'', <arg n=``2''>, <arg n=``3''>, <arg n = ``4''> and <arg m> represent meta tags, which add or modify information to Agent or Patient, we identify this information using the tag ``Other'' and <rel> indicates the relation or the action present in the sentence, we identify this information using the tag ``Relation''. 

\subsubsection{Feature extraction}

To annotate the data, we take each sentence of the Propbank annotated sentence, and then parse the sentences based on the rules generated in the previous step %(the parsing is done via the Earley Parsing algorithm \cite{earley1970}). 

\begin{figure}
 \centering
 \includegraphics[width=0.20\textwidth,natwidth=610,natheight=642]{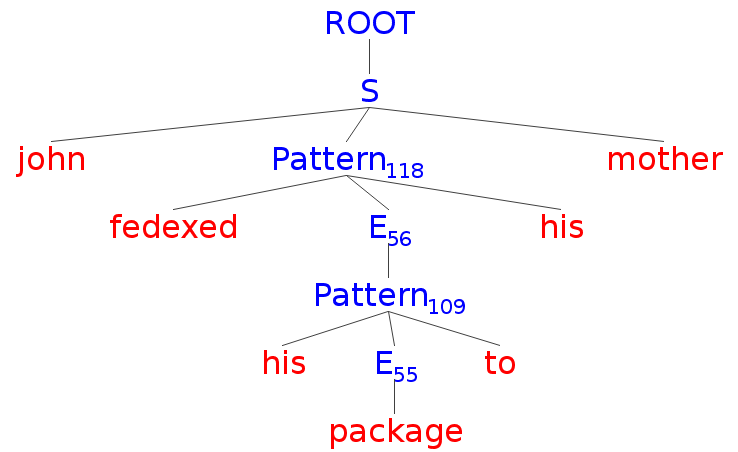}
 \caption{Parsing using grammar rules learnt from raw text using m-ADIOS}
 \label{fig:rawParseSentenceExample}
 
\end{figure}

\begin{figure}
 
 \centering
 \includegraphics[width=0.20\textwidth,natwidth=610,natheight=642]{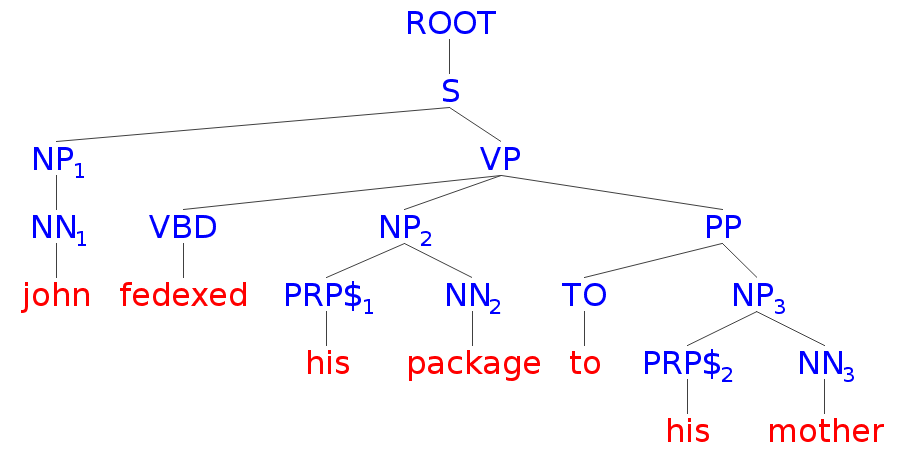}
 
 \caption{Parsing using Stanford Parser}
 \label{fig:stanfordParse}
\end{figure}

For illustration purposes let\'s look at the sentence ``John FedExed his package to his mother''. This sentence when parsed gives us a tree shown in Figure \ref{fig:rawParseSentenceExample}. The syntactic tree using rules derived from language deviates from the parse done using Stanford parser \cite{socher2013} (see Figure \ref{fig:stanfordParse}) which is considered a state-of-art parser. Even though it does not match the output of a standard parser, rules learned from language consistently parse similar sentences the same way. This has the advantage that - we can use the rules to train model for identifying semantic roles. 

Once we parsed the sentence,  we extract various features of the parse tree and use them as our features for the classification. The features we have used are shown in Table \ref{table:rawFeaturesForClassification}. Two words before and after the pattern are used to capture the context of the pattern. Pattern labeling is analogous to the head word, words inside the pattern are analogous to the path between the semantic roles, and length of the pattern gives information about the span of the pattern. The use of surface features of the parse tree to identify semantic roles is a standard practice in several supervised and unsupervised methodologies of semantic role labeling. %\cite{gildea2002b,gildea2002c,giuglea2006,pradhan2005a,lang2010,lang2011a}.

\begin{table}
\centering
 \caption{Features for classification}
 \begin{tabular}{l}
 \hline
 \textbf{ Features used for classification}\\ \hline
 1. Head of Phrase \\
 2. Word two words before the pattern \\
 3. Word one word before the pattern \\
 4. Words inside the pattern \\
 5. Word adjacent to the pattern \\
 6. Word two words after the pattern \\
 7. Length of the Pattern \\
 \hline
 \end{tabular}
 
 \label{table:rawFeaturesForClassification}
 \end{table}

\subsubsection{Labeling the data}
Since patterns and equivalent classes learned using the m-ADIOS algorithm does not have control over phrase boundaries, there are cases in which multiple semantic roles can be encompassed inside a pattern. For example, 
Figure \ref{fig:DatasetCreation} shows that ``Pattern\_118'' encodes ``Patient'' and partially encodes ``Other''. For simplifying the task, we assume that we encapsulate the semantic role even if we encapsulate a single word inside the semantic role. So we give the label ``Patient\_Other'' for ``Pattern\_118''. 
 
%  \begin{figure}
%  \centering
%  \includegraphics[width=0.5\textwidth]{LabelCreation}
%  
%  \caption{Label creation example}
%  \label{fig:labelCreation}
% \end{figure}
 In this study we focused on identifying agent, patient and relation, we simplify our class labels by considering them as a triplet of boolean decisions. The triplet is made of ``<Agent, Patient, Relation>'', so the class label for ``Pattern\_118'' would be ``<false,true,false>'' . The classification task using the above triplet ``<Agent, Patient, Relation>'' is an 8 class classification with possibility of each entity being true or false.

 Based on seven features shown in Table \ref{table:rawFeaturesForClassification}, and the class label given to each instance of the patterns, we do a classification via a 10-fold cross validation using Bayes, Naive Bayes, and Random Forest classifiers using Weka \cite{hall2009}. 

 \subsection{Summary of Dataset Creation} \label{sec:methodologyForRawData}

To test the performance of m-ADIOS in the generating structure which encodes semantic roles, we use sentences in PropBank and parsed sentences obtained after parsing sentences using rules learned from the language. Let us look at the example closely, the sentence ``john fedexed his package to his mother'', is parsed using the rules generated from the PropBank corpus. After parsing, we end up with a tree like structure, which is represented using a bracketing scheme. 

\begin{figure}
 \centering
 \includegraphics[width=0.50\textwidth]{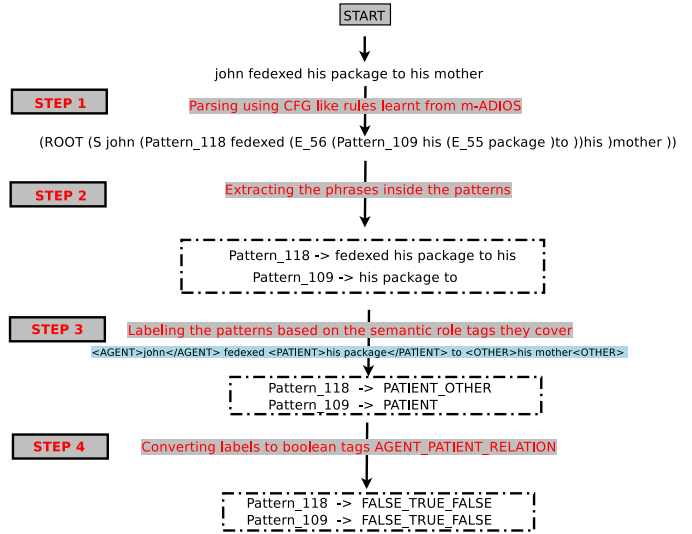}
 \caption{Creating dataset for classification}
 \label{fig:DatasetCreation}
\end{figure}
Figure \ref{fig:rawParseSentenceExample} shows tree representation of the same bracketing scheme. Figure \ref{fig:stanfordParse} represents the structure using the Stanford parser. We can observe that patterns take the role of ``head'' similar to that of Noun Phrase (``NP''), Verb Phrase (``VP'') etc. Patterns learnt from the grammar rules do not have correct phrase boundaries in contrast to the strict adherence to phrase boundaries in traditional parsers (Figure \ref{fig:stanfordParse}). Since PropBank follows strict phrase boundaries, which adheres to Penn-tree bank annotations, our patterns transgress the defined phrase boundaries. As a result of these transgressions we some times have more than one role inside the pattern.
 
In step 3 in the Figure \ref{fig:DatasetCreation} we capture the semantic roles encapsulated by the patterns. Here we can see that ``Pattern\_118'' encodes ``Patient'' and partially encodes ``Other.'' For simplifying the task, we assume that we encapsulate the semantic role even if we encapsulate a single word inside the semantic role. So we give the label ``Patient\_Other'' for ``Pattern\_118''.
 
We perform these four steps on all the sentences for which we have learned rules using the m-ADIOS. The classification task using the above triplet ``<Agent, Patient, Relation>'' is an 8 class classification with a possibility of each entity being true or false. To summarize, given a pattern we would predict whether it encodes an agent or a patient or a relation or any combination of them. 

 %Inspired by features used in the supervised models for SRL, such as path between agent and patient, head word, active or passive sentence, part-of-speech of the words in the path, and context around the head word, we extracted similar features from our parse tree obtained using grammar rules learned from data. The feature are shown in Table \ref{table:rawFeaturesForClassification}. The dataset consists of $575$ different patterns and a total $1309$ instances of patterns. Since on an average there are only $2.3$ instances per pattern, we can expect to get high accuracy as the patterns obtained do not cover many sentences, and when trained for a single instance of the pattern itself would give an accuracy of $\sim 50\%$.

\subsection{Results and Discussion}

Based on the 7 features shown in Table \ref{table:rawFeaturesForClassification}, and the class label given to each instance of the patterns following Section \ref {sec:methodologyForRawData}, we did a classification via a 10-fold cross validation using Bayes, Naive Bayes and Random Forest models. We report all the results using the measures precision $P=\frac{\mathit{No.\ of\ correct\ constituents\ identified}}{\mathit{Toal\ no.\ of\ constituents\ identified}}$; recall $R=\frac{\mathit{No.\ of\ correct\ constituents\ identified}}{\mathit{Total\ no.\ of\ correct\ constituents}}$; f-measure $F=\frac{2\times{P}{R}}{P+R}$ and inter-rater reliability using Cohen\'s kappa $K$.

Table \ref{table:rawWordsResults} shows the classification results. As expected the accuracy is reasonably high given that the chance of finding the correct tag is only $0.125$. The classification is done with a great amount of confidence as well (k > $0.6$). These can be attributed to the sparsity in the patterns obtained from the language. 

\begin{table}
\centering
\caption{Results showing the classification for actual sentences}
 \begin{tabular}{c c c c c}
 \hline
Classifier&P&R&F&K \\ \hline
BayesNet&0.81&0.795&0.789&0.711 \\ 
NaiveBayes &0.807&0.767&0.749& 0.664 \\ 
Random Forest&0.79&0.75&0.75&0.64\\
\hline
\end{tabular}

\label{table:rawWordsResults}
\end{table}

\begin{table}
\centering
\caption{Results of state-of-art unsupervised semantic role labeling models}
 \begin{tabular}{c c c c c}
 \hline
Work&Purity&Precision&F1 \\ \hline
\cite{lang2010}&0.80&0.77&0.78\\
\cite{lang2011a}&0.89&0.73&0.80\\
\hline
\end{tabular}

\label{table:baselineResults}
\end{table}

 Semantic roles are defined by order and kind of words present in the sentence, gave us the motivation to explore the distributional information around the words, and hierarchical relations among them. We further developed the m-ADIOS algorithm inspired from ADIOS \cite{solan2005}. In the m-ADIOS algorithm, the structural information learned is strictly dependent on the left and the right context in which words appear. We generalized this by assuming that only when two patterns have a perfect overlap on the right and left context or when there is a match with a left or right context of the two patterns and a significant overlap in their equivalent structures thereby reducing the ambiguity in the grammar.
 
 \cite{lang2010,lang2011a} used dependency trees from PropBank to extract features like predicate-lemma, argument-lemma, argument part-of-speech, and part-of-speech of left and rightmost child of argument to cluster the semantic roles in the sentences. Table \ref{table:baselineResults} shows the baseline accuracies of current state-of-art unsupervised srl models. Even though their accuracy is high, their latent dependence of human annotated data {dependency parsers, part-of-speech taggers} makes it pseudo-unsupervised. Several other unsupervised models of semantic role labeling (see Table \ref{table:literature_survey}) also share the same issues.
 
 Our results show that we can learn structure from language, in a data-driven fashion, and with little training we were able to identify the presence and absence of semantic roles. Even though the precision and recall scores may be lower when compared to existing unsupervised models, by not being dependent on human annotated data the proposed method would help to identify semantic roles for languages for which syntactic parsers developed from human annotated data are unavailable.
 
\subsection{Future Work} 
The strict generalization in our model allows patterns that share a strong overlap in their context to be termed equivalent. After learning the rules, we convert them to grammar rules and parse the sentences from which we have learned the rules. In theory, these rules can be extended to any other sentences, which have the same words. But in practice this becomes a major drawback as we have the problem of sparsity simply because many words can occur in different contexts, and the order they appear in makes a big difference on the forming of patterns. To overcome this problem we plan to the method used by \cite{datla2014} to induce pseudo parts of speech, and learn structures based on the induces parts of speech.

In future we would like to extend our model for other languages, and also to the languages that are resource constraint.

 \bibliographystyle{aaai}
 \bibliography{references-dissertation}

\end{document}